\documentclass{article}
\usepackage{spconf,graphicx}
\usepackage{multirow}
\usepackage{subcaption}
\usepackage{booktabs}
\usepackage{amssymb}
\usepackage{latexsym}
\title{FedPartWhole: Federated domain generalization via consistent part-whole hierarchies}
\name{Ahmed Radwan, Mohamed S. Shehata}
\address{University of British Columbia\\
Department of Computer Science\\
3333 University Way, Kelowna, BC
}
\begin{document}
%\ninept
%
\maketitle
\begin{abstract}
Federated Domain Generalization (FedDG), aims to tackle the challenge of generalizing to unseen domains at test time while catering to the data privacy constraints that prevent centralized data storage from different domains originating at various clients. Existing approaches can be broadly categorized into four groups: domain alignment, data manipulation, learning strategies, and optimization of model aggregation weights. This paper proposes a novel approach to Federated Domain Generalization that tackles the problem from the perspective of the backbone model architecture. The core principle is that objects, even under substantial domain shifts and appearance variations, maintain a consistent hierarchical structure of parts and wholes. For instance, a photograph and a sketch of a dog share the same hierarchical organization, consisting of a head, body, limbs, and so on. The introduced architecture explicitly incorporates a feature representation for the image parse tree. To the best of our knowledge, this is the first work to tackle Federated Domain Generalization from a model architecture standpoint. Our approach outperforms a convolutional architecture of comparable size by over 12\%, despite utilizing fewer parameters. Additionally, it is inherently interpretable, contrary to the black-box nature of CNNs, which fosters trust in its predictions, a crucial asset in federated learning.
\end{abstract}
%


\begin{figure}[h]
  \centering
   \includegraphics[width=\linewidth]{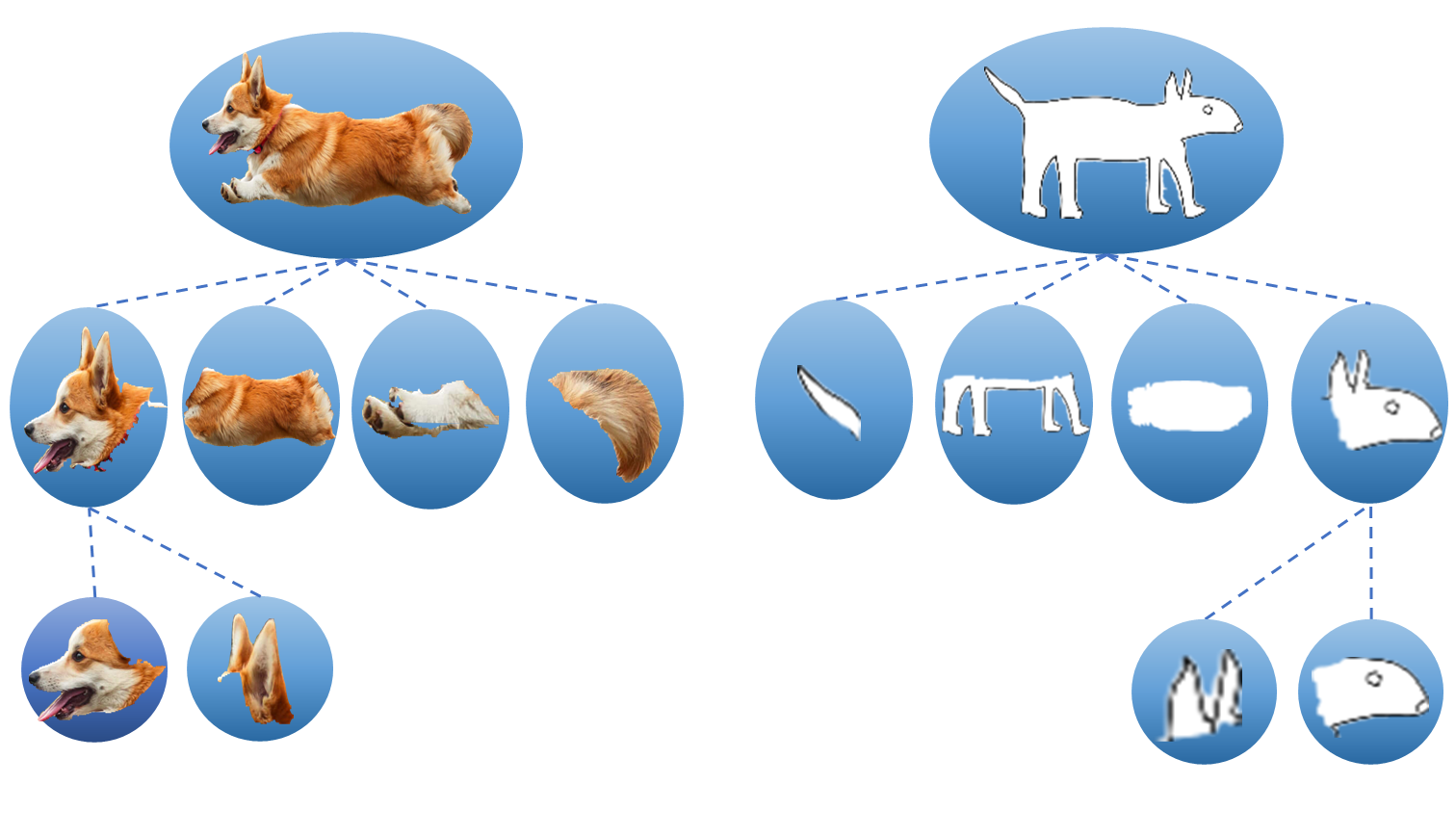}

   \caption{The Figure illustrates the motivation behind our approach. Objects will maintain their compositional hierarchical structure despite any change in appearance, which aids domain generalization. On the left, the parse tree of a picture of a dog, while the right-hand side depicts the parse tree for a sketch drawing of a dog. The figure illustrates that both images have the same parse tree, despite the huge appearance discrepancy.}
   \label{fig:hierarchy}
\end{figure}
\section{Introduction}
\label{intro}

%review: add to fig2 input image +

\begin{figure*}[h]
  \centering
   \includegraphics[width=\linewidth]{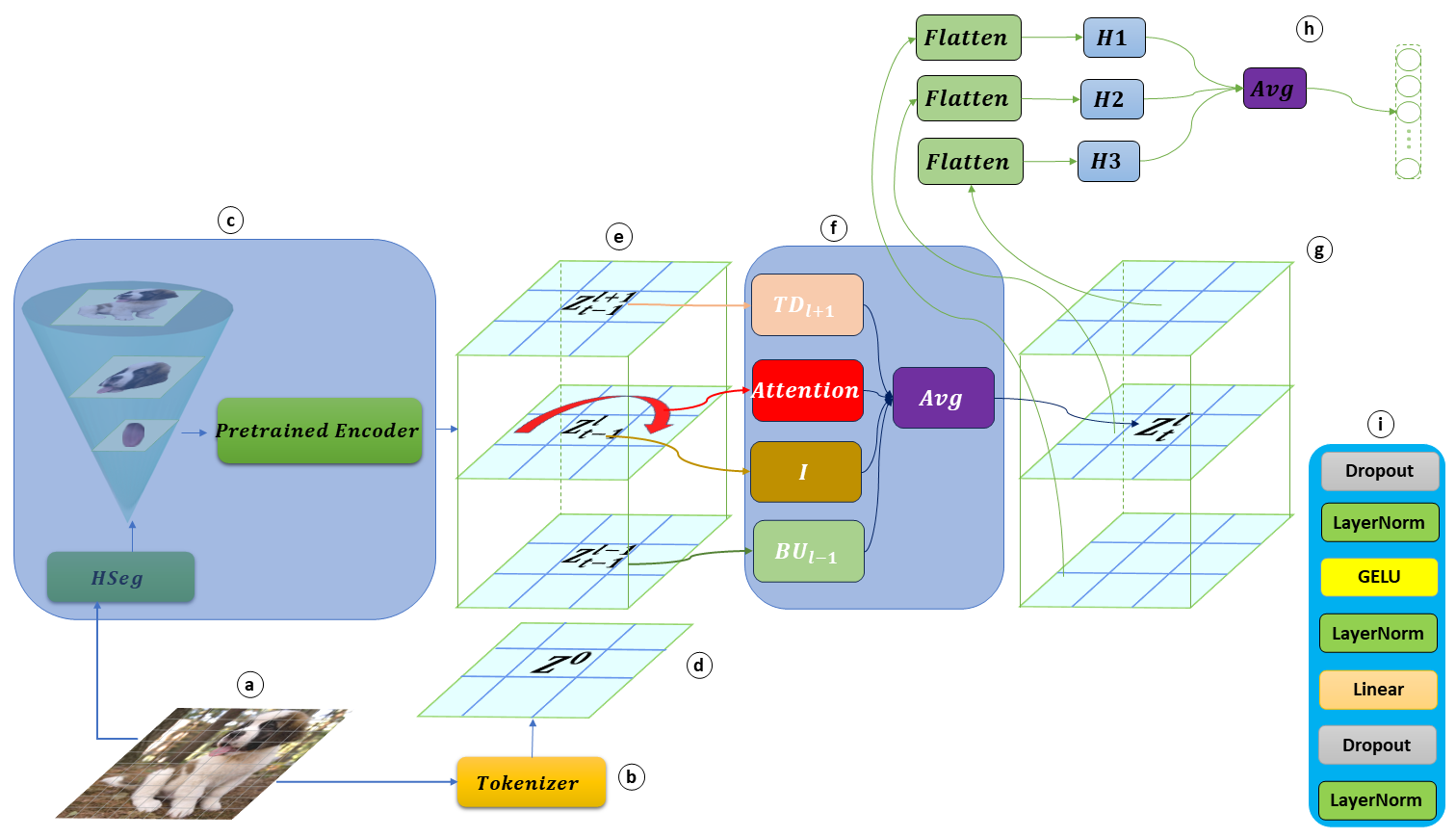}
  \caption{The Figure illustrates the  CCNet architecture. a) the input image. b) the input tokenizer. c) the initialization pipeline. d) the tokenized input e) the hidden representation at t = 0. f) The network architecture. g) the hidden representation at t = 1. h) the softmax classification output. i) the structure of the classification head. }
  \label{fig:arch}
\end{figure*}

Generalization, the ability to apply learned concepts to unseen situations, is a fundamental challenge in achieving human-level artificial intelligence. Machine learning algorithms have achieved remarkable success under the assumption of identically and independently distributed (i.i.d.) data, even surpassing human level performance \cite{he2015delving} on popular benchmarks like ImageNet \cite{deng2009imagenet}. However, their performance often degrades significantly when faced with domain shifts in the test data \cite{walsh2022automated}. This challenge has motivated extensive research on models that can generalize to unseen domains during testing. The work in \cite{blanchard2011generalizing} was the earliest to propose a method to tackle this problem, which is known as the Domain Generalization (DG) problem. Popular DG approaches rely on exposing models to data from multiple source domains during the training time. This allows the model to learn features that are invariant across these domains, ideally leading to robustness against domain shifts at test time.
However, this relies on a centralized training setup where all source domains reside on a single server, enabling the model to learn from data from different domains within a single mini-batch \cite{dou2019domain} \cite{li2018domain} \cite{sun2016deep}. This approach presents practical limitations. In real-world scenarios, data from different domains often originates from distributed devices or organizations, making centralized storage infeasible. Additionally, data privacy regulations and laws, such as GDPR \cite{truong2021privacy}, may restrict the sharing of raw data across entities.

Federated Domain Generalization (FedDG) tackles this challenge by enabling models to learn from distributed data while preserving privacy. Existing FedDG approaches can be broadly categorized into four main areas: 1) Federated Domain Alignment, which focuses on aligning features from different domains to reduce domain discrepancies. 2) Data Manipulation, which aims to improve model performance by augmenting existing source domains or generating new synthetic domains. 3) Learning Strategies, which explore novel training algorithms specifically designed for the FedDG setting. 4) Aggregation Optimization, which focuses on developing techniques for effectively aggregating model parameters and gradients received from participating devices.

Pioneering work done in \cite{hinton2023represent} introduced a theoretical framework called GLOM, which is a proposal of how neural networks could learn to represent part-whole hierarchies. \cite{garau2022interpretable} implemented the framework, dubbed as Agglomerator, and showed promising results on toy datasets. While \cite{radwan2023distilling}, showed how to pre-train the network effectively to work on more complex natural scenes, benefitting from the pre-trained knowledge of other commonly used architectures.

In this paper, we build upon the theoretical framework of GLOM and propose a novel architecture, a Cortical Columns Network (CCNet), specifically designed to enhance generalization capabilities and tackle the FedDG problem. The main motivation behind this approach is the observation that objects across different domains belonging to the same class share a consistent hierarchical structure of parts and wholes, Figure \ref{fig:hierarchy} illustrates the concept. The proposed architecture explicitly models the part-whole hierarchies in its feature maps. Facilitating the classification decision based on the recognized parse tree. Our main contributions can be summarized as follows:
%review: check consistency in writing of the word (backbone architecture)
\begin{itemize}
    \item We present a novel lightweight backbone architecture, CCNet. This architecture explicitly models the compositional structure of visual objects within its feature map. To the best of our knowledge, this is the first paper to address the FedDG problem from a backbone architectural point of view.
    
    \item The proposed backbone can be used with existing FedDG algorithms, making our development orthogonal to the existing approaches. When used as a backbone in different FedDG algorithms, the results show strong performance improvements over the widely established CNN architecture, with fewer parameters.
    
    \item Contrary to many existing backbones, the proposed backbone architecture does not require a computationally expensive pre-training step thanks to its explicit representation of the compositional structure of the objects.
    
\end{itemize}

\section{Related Work}
\label{sec:related}

\subsection{Federated Domain Generelization}
Federated domain generalization (FDG) aims to tackle both the domain shift and data privacy problem simultaneously. The research efforts on FDG can be mainly categorized into 4 main groups. 1) \textbf{\textit{Domain alignment}}, aims to align the representations of multiple source domains, which helps the model mitigate domain shifts and generalize to unseen domains at test time. Adversarial domain alignment, proposed by \cite{peng2019federated}, involves training a discriminator to identify the feature distribution of the source domain, while a generator learns to produce robust features to confuse the discriminator. While \cite{chen2022federated} proposed an approach to minimize the maximum mean discrepancy between source and target domains. 2) \textbf{\textit{Data Manipulation}}, based techniques, aim to diversify and enlarge the source domains as much as possible by manipulating the available data. \cite{shenaj2023learning} applied style transfer augmentation to mimic the target domain styles within the source images. While \cite{liu2021feddg} uses a powerful augmentation based on the frequency spectrum information. 3) \textbf{\textit{Learning strategies}}, encompasses a variety of techniques and algorithms that aim to learn a robust global model. Techniques include channel decoupling \cite{shen2022cd2}, and regularization strategies \cite{huang2020self}. 4) \textbf{\textit{Aggregation optimization}}, techniques aim to find optimal ways to aggregate model parameters or gradient updates. \cite{alekseenko2024distance} proposed using distance metrics to decrease the influence of diverging clients. While \cite{tian2021privacy} optimizes gradient updates by introducing a loss on the gradient alignment.
Finally, our work represents both a novel research direction and orthogonal to the existing directions in the realm of FedDG. We tackle the problem by proposing an architecture that explicitly models the scene's parse tree, which should be invariant under domain shifts.

\subsection{Related Foundational Models}
Few endeavors have been made to develop architectures capable of representing scene parse trees. Our work falls in line with such models.

\subsubsection{Capsule Networks}
One such architecture is the Capsule Network \cite{sabour2017dynamic}. Capsule Networks create layers of capsules, which are fundamental elements responsible for detecting objects or their constituent parts. These capsules, employing dynamic routing by agreement, establish connections with their corresponding parent capsules, effectively forming a parse tree of the scene. While Capsule Networks have demonstrated competitive performance on easier datasets, although they face challenges in matching the performance of convolutional or transformer-based approaches on more intricate datasets.

\subsubsection{Agglomerator}
The theoretical framework of GLOM was introduced \cite{hinton2023represent} to remedy some of the issues with capsule networks. Agglomerator \cite{garau2022interpretable} is an implementation of GLOM. Agglomerator's authors proposed training to undergo two distinct training phases. Firstly, it is pre-trained using the supervised contrastive loss. Subsequently, the network is trained with the cross entropy loss. Agglomerator alleviated some of the issues with capsule networks. However, \cite{radwan2023distilling} showed that it fails to deal with more complex natural images beyond toy datasets, such as those in ImageNet \cite{deng2009imagenet}. To improve the performance, and pre-train the network efficiently, they proposed a knowledge-distillation approach to pre-train Agglomerator using the available knowledge in other pre-trained models.

\subsubsection{Segment Anything}
The Segment Anything Model, SAM, was recently introduced as a new foundation model for image segmentation on images from any domain. Segmentation is one of the most fundamental computer vision tasks and finds its way to many applications. A segmentation foundation model, that offers zero-shot generalization capabilities without requiring domain-specific training, is very likely to have beneficial information. SAM was trained on more than a billion segmentation masks, which explains its performance. SAM can be prompted to generate segmentation masks using either a bounding box, text description, a mask, a point prompt, or a combination of the prompts. Since some prompts could be ambiguous, SAM offers a multi-output segmentation mask to resolve the ambiguity. In our work, we leverage the domain generalized knowledge in SAM and use it to generate an initial scene parse tree. We use the embeddings of the generated parse tree to initialize the CCNet feature map.

\begin{figure}[]
    \centering    \includegraphics[width=\linewidth]{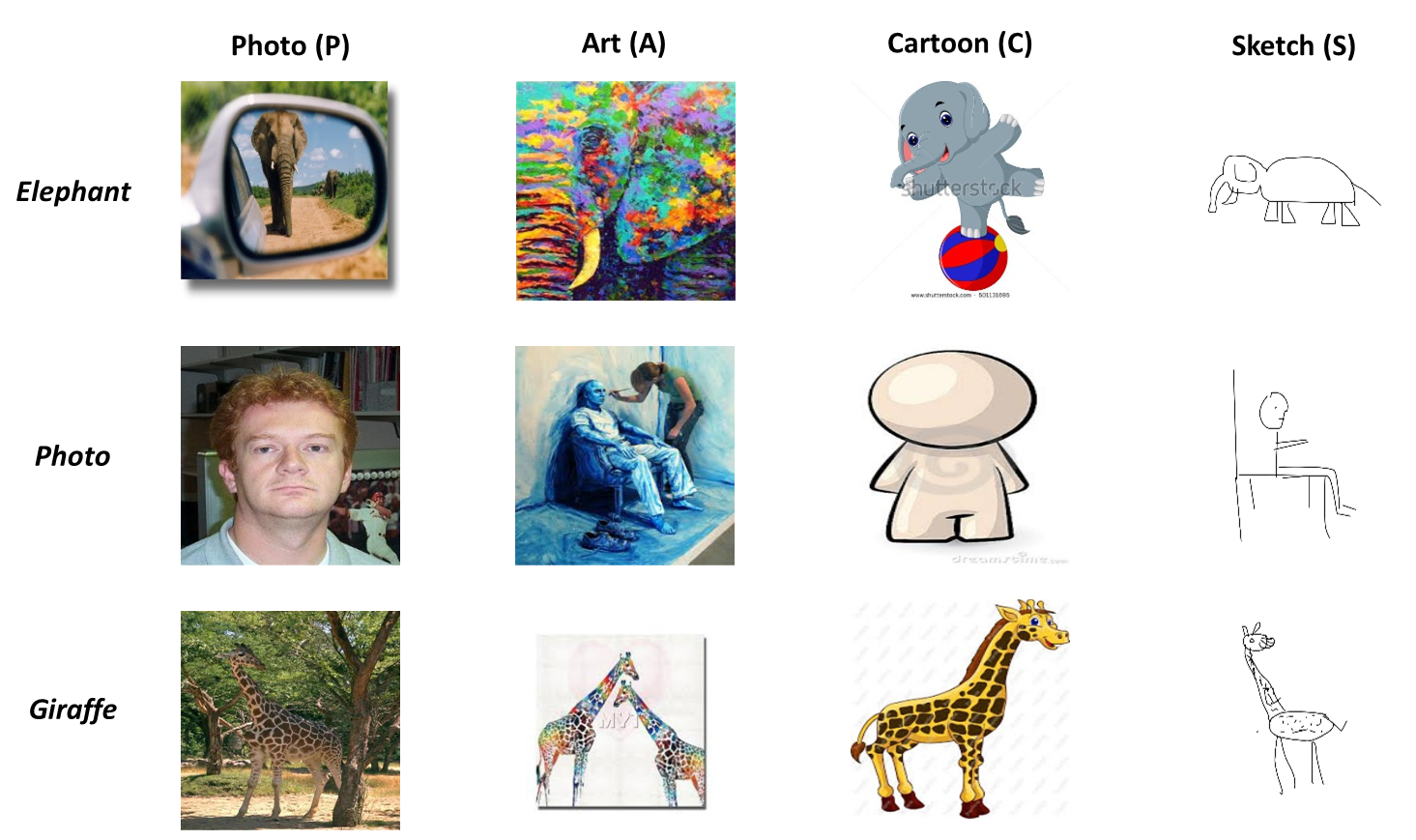}
    \caption{The Figure depicts a sample of the PACS dataset from the four different domains, illustrating the challenge of domain generalization under significant domain shift.}
    \label{fig:pacs_sample}
\end{figure}

\section{Proposed Method}

In this section, we describe our proposed system, inspired by the column-like arrangement of neurons found in the human's neocortex and by the theoretical framework, GLOM. We will start by introducing the main building blocks of the proposed architecture, dubbed CCNet. Firstly, the input image is passed through a convolutional network to obtain $N = n \times n$ feature map patches which will serve as the input to our system. The system represents the part-whole hierarchy of the image using columns. Each Column, $C_i$, specializes in processing a single spatial location and is assigned to process the corresponding $i^th$ input patch, with $i \in \{1,\dots,N\}$. A Column comprises $L$ levels of embeddings. An embedding vector at a time step $t$, and level $l$ of the $i^th$ C is denoted by $Z_t^{l,i}$, where $l \in \{1,\dots,L\}$. 

The latent representation of the network, $Z$, is an explicit modeling of the part-whole hierarchy of objects and their parts in a visual scene. Each embedding level in the column is a latent representation of the image patch at some level of abstraction. The lowest level of the column represents the low-level information present in the patch and as we go towards the highest level we expect to see a vector representing the object of which the patch is a part of till we arrive at a scene-level representation.

\subsection{Input Tokenization}

The first step in our architecture is transforming the input image into patches of input tokens. This is achieved by firstly passing the image through a convolutional network that transforms an input image into $ n \times n$ feature map. Each of the feature map patches will be embedded into the lowest level of its corresponding column, where the $Z_0^{0,i}$ vector will come from the $i^th$ feature map patch.

%review add the keyword fed average and move the table position

% \usepackage{graphicx}
\begin{table*}[]
\caption{Leave-one-domain-out results on both Domain Generalization datasets (PACS and VLCS) comparing CCNet and CNN as backbones for different algorithms}
\centering
\resizebox{\textwidth}{!}{%
\begin{tabular}{llllll|lllll}
\hline
                         & \multicolumn{5}{c|}{PACS}                                                     & \multicolumn{5}{c|}{VLCS}                                                     \\ \hline
Method                   & P       & A       & C       & \multicolumn{1}{l|}{S}       & Avg              & V       & L       & C       & \multicolumn{1}{l|}{S}       & AVG              \\ \hline
FedAvg + CNN             & 76.05\% & 43.80\% & 52.39\% & \multicolumn{1}{l|}{47.67\%} & 54.98\%          & 58.24\% & 56.46\% & 90.85\% & \multicolumn{1}{l|}{65.18\%} & 67.68\%          \\
FedAvg + CCNet (ours)   & 91.92\% & 75.10\% & 65.10\% & \multicolumn{1}{l|}{70.78\%} & \textbf{75.73\%} & 66.24\% & 56.71\% & 93.63\% & \multicolumn{1}{l|}{70.36\%} & \textbf{71.74\%} \\ \hline
AM + CNN                 & 69.52\% & 57.18\% & 64.89\% & \multicolumn{1}{l|}{74.73\%} & 66.58\%          & 53.11\% & 60.98\% & 78.07\% & \multicolumn{1}{l|}{60.81\%} & 63.24\%          \\
AM + CCNet (ours)       & 90.84\% & 74.02\% & 66.13\% & \multicolumn{1}{l|}{66.35\%} & \textbf{74.34\%} & 64.96\% & 54.58\% & 93.16\% & \multicolumn{1}{l|}{64.48\%} & \textbf{69.30\%} \\ \hline
FedProx + CNN            & 73.35\% & 35.55\% & 53.11\% & \multicolumn{1}{l|}{28.76\%} & 47.69\%          & 56.27\% & 59.85\% & 85.14\% & \multicolumn{1}{l|}{61.12\%} & 65.60\%          \\
FedProx + CCNet (ours)  & 92.16\% & 77.29\% & 67.15\% & \multicolumn{1}{l|}{67.63\%} & \textbf{76.06\%} & 68.11\% & 56.84\% & 94.58\% & \multicolumn{1}{l|}{68.02\%} & \textbf{71.89\%} \\ \hline
RSC + CNN                & 74.79\% & 62.60\% & 70.05\% & \multicolumn{1}{l|}{73.96\%} & 70.35\%          & 60.51\% & 61.10\% & 91.27\% & \multicolumn{1}{l|}{68.02\%} & 70.23\%          \\
RSC + CCNet (ours)      & 90.48\% & 75.44\% & 66.08\% & \multicolumn{1}{l|}{65.77\%} & \textbf{74.44\%} & 68.31\% & 56.71\% & 94.10\% & \multicolumn{1}{l|}{66.80\%} & \textbf{71.48\%} \\ \hline
Scaffold + CNN           & 74.43\% & 38.09\% & 56.83\% & \multicolumn{1}{l|}{29.02\%} & 49.59\%          & 57.16\% & 60.23\% & 88.92\% & \multicolumn{1}{l|}{61.83\%} & 67.04\%          \\
Scaffold + CCNet (ours) & 91.02\% & 75.20\% & 66.60\% & \multicolumn{1}{l|}{66.40\%} & \textbf{74.81\%} & 65.15\% & 57.09\% & 93.40\% & \multicolumn{1}{l|}{64.97\%} & \textbf{70.15\%} \\ \hline
\end{tabular}%
}
\label{tab:algos}
\end{table*}

\subsection{CCNet}
CCNet consists of 4 main modules that are responsible for processing the hidden representation at a time step t to obtain the representation at the next time step, $Z_{t+1}$. Firstly, the bottom-up (BU) module predicts the next level in a particular column from the previous level in the same column in the previous time step. The BU module is a stack of $L$ MLP networks, where each two consecutive levels are connected by an MLP network. Formally, the $l^th$ MLP in the BU module predicts $Z^{l,i}_{t+1}$ from $Z^{l-1,i}_t$. Similarly, we have the Top Down (TD) module, which predicts a lower level from a high level in a particular column in the previous time step. The TD module consists of $l-1$ MLP networks, where the $l^th$ network predicts $Z^{l,i}_{t+1}$ from $Z^{l+1,i}_t$. The Identity ($I$) module simply copies the value of a level in a column to the same level in the next time step. Lastly, since each column represents a single patch from the image, a simple Attention module is needed to aggregate spatial information from neighboring columns. A vector obtains its attention weight from a weighted sum across the neighboring columns at the same level, $l$. Formally, the attention weight given at a location $i$ to a vector $Z^{l,j}$  is computed as

\begin{equation}   
\frac{e^{\beta Z^{l,i}.Z^{l,j}}}{\sum_{k}^{K}e^{\beta Z^{l,i}.Z^{l,k}}}
\end{equation}   

Where $K$ is the set of all neighboring locations that $i$ attends to, $\beta$ is a parameter that acts like an inverse temperature to adjust the attention sharpness. 
Finally, contributions from the 4 modules are averaged with learnable weights to obtain the representation for a particular level vector.

\subsection{Training}
We train CCNet in a supervised manner on the source domains using the cross entropy classification loss. We perform a single time step update from $t=0$ to $t=1$ as illustrated in Figure \ref{fig:init} to obtain the hierarchical representation onto which we build the 3 classification heads. Instead of a random initialization of the feature map at $t=0$, Segment Anything Model (SAM) is utilized to obtain a hierarchical segmentation of the scene which is encoded and used as an initialization for the hidden representation. While SAM doesn't directly offer a hierarchical segmentation of the scene, the next subsection will explain the details of the process.

\subsection{Generating SAM Masks}
\label{subsec:sam_masks_gen}
We capitalize on the fact that SAM can offer multi-level segmentation masks in its output to disambiguate an ambiguous point prompt. To obtain the initial hidden representation, $Z_0$, We start by dividing the image into several patches consistent with the discussed Input Convolutional Tokenizer module. For each patch, we use the middle point as an input prompt for SAM. SAM will return 3 level segmentation masks where a mask typically represents the whole object, a mask representing a part of the object, and finally a mask representing a sub-part of the object. More concretely, for each image in a training batch of size $B$ we divide the image into $N$ patches, and generate $N$ prompts points corresponding to the center points of each image patch as an input to SAM which in response will generate $L$ outputs per point prompt, each output is a binary mask of dimensions $H \times W$, where $H$, and $W$ are the input image dimensions. The output from SAM is finally given by the 5D tensor $M \in \mathbb{R}^{B \times N \times L \times H \times W}$.

\subsection{Initializing the Feature Maps}
We harness a downsized variant of the pre-trained MaxVIT \cite{tu2022maxvit} image encoder, referred to as pico MaxVIT, to generate embeddings for the obtained SAM masks. The masks $M$ are input to the MaxVIT encoder, producing embedding representations for $M$, denoted as $M_e \in \mathbb{R}^{B \times N \times L \times D_e}$. Here, $D_e$ corresponds to the image embedding size that aligns with the previously mentioned hidden representation size in $Z_e$. We set $Z_0 = M_e$, providing a strong prior on the representations CCNet should obtain for the image. It is worth noting that this initialization can be error-prone, and the network is supposed to learn normally from clients data

 \subsection{Visualization}
\begin{figure}
  \centering
   \includegraphics[width=\linewidth]{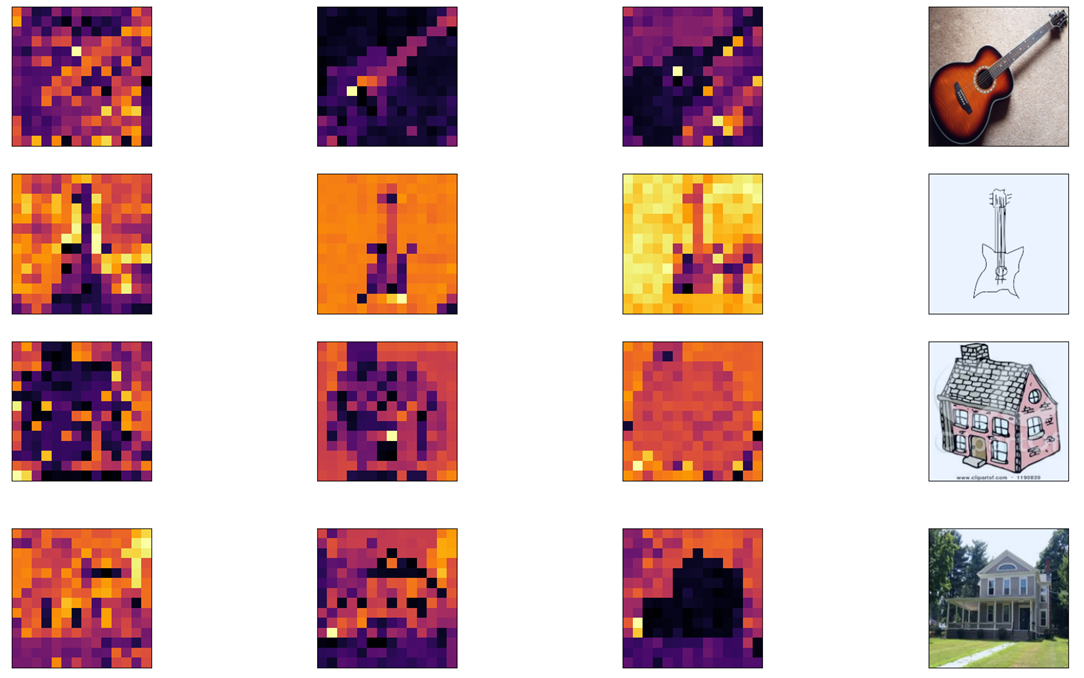}
  \caption{The Figure illustrates the islands of agreements formed at $t = 0$ in CCNet. The right most columns show higher levels in the hirer achy, with the last column being the input image. The Figure illustrates a similar hierarchical structure of images belonging to different domains. }
  \label{fig:islands}
\end{figure}

Following the approach of the original work on Agglomerator, we have carried out a visual analysis of the islands of agreement across multiple images. The results indicate that the model has effectively captured meaningful hierarchical representations of the objects within the scenes. This is evident in the embedding vectors, which display strong similarities across locations that correspond to the same object or its components. Figure \ref{fig:islands} illustrates these islands of agreement at various levels. Patches with resembling embedding vectors are depicted using similar colours. Notably, as we progress up the hierarchy, towards the rightmost columns, we observe the convergence of embedding vectors into two distinct 'islands.' One represents the foreground elements, and the other corresponds to the background.

\section{Experiments}

% 
%\begin{table}[]
%\resizebox{\columnwidth}{!}{
%{\renewcommand{\arraystretch}{2}
%\begin{tabular}{@{}ccccccc@{}}
%\toprule
%                                &                              %              & \multicolumn{4}{c}{PACS}                       %                                      &                 \\ %\midrule
%\multicolumn{1}{c|}{Params (M)} & \multicolumn{1}{c|}{Method}  %              & P              & A             & C             %& \multicolumn{1}{c|}{S}              & Avg             \\ %\midrule
%\multicolumn{1}{c|}{4.2}        & \multicolumn{1}{c|}%{MobileNet \cite{howard2017mobilenets}}             & 76.05          & 43.8          & 52.39         & \multicolumn{1}{c|}{47.67}          & 54.98           \\
%\multicolumn{1}{c|}{3}          & \multicolumn{1}{c|}{Distill Part Whole \cite{radwan2023distilling}}    & 31.5           & 24.46         & 30.84         & \multicolumn{1}{c|}{32.45}          & 29.81           \\
%\multicolumn{1}{c|}{3}          & \multicolumn{1}{c|}{Agglomerator \cite{garau2022interpretable}}          & 39.34          & 26.17         & 35.03         & \multicolumn{1}{c|}{38.71}          & 34.81           \\
%\multicolumn{1}{c|}{4}          & \multicolumn{1}{c|}{CCNet (ours)} & \textbf{91.92} & \textbf{75.1} & \textbf{65.1} & \multicolumn{1}{c|}{\textbf{70.78}} & \textbf{75.725} \\ \bottomrule
%\end{tabular}}}
%\caption{Leave-one-domain-out results on PACS dataset.}
%\label{tab:pacs}
%\end{table}

\begin{table*}[]
\caption{Leave-one-domain-out related backbones results on both PACS and VLCS datasets using FedAvg method}
\centering
\resizebox{\textwidth}{!}{%
\begin{tabular}{@{}cccccccccccc@{}}
\toprule
                                &                                            & \multicolumn{5}{c}{PACS}                                                                                                    & \multicolumn{5}{c}{VLCS}                                                                                                              \\ \midrule
\multicolumn{1}{c|}{Params (M)} & \multicolumn{1}{c|}{Method}                & P              & A             & C             & \multicolumn{1}{c|}{S}              & \multicolumn{1}{c|}{Avg}             & \multicolumn{1}{c}{V} & \multicolumn{1}{c}{L} & \multicolumn{1}{c}{C} & \multicolumn{1}{c|}{S}              & \multicolumn{1}{c}{Avg} \\ \midrule
\multicolumn{1}{c|}{3}          & \multicolumn{1}{c|}{Distill Part Whole \cite{radwan2023distilling}}    & 31.5           & 24.46         & 30.84         & \multicolumn{1}{c|}{32.45}          & \multicolumn{1}{c|}{29.81}           & 46.2                  & 50.82                 & 63.68                 & \multicolumn{1}{l|}{43.55}          & 51.06                   \\
\multicolumn{1}{c|}{3}          & \multicolumn{1}{c|}{Agglomerator \cite{garau2022interpretable}}          & 39.34          & 26.17         & 35.03         & \multicolumn{1}{c|}{38.71}          & \multicolumn{1}{c|}{34.81}           & 46.3                  & 52.45                 & 64.15                 & \multicolumn{1}{l|}{45.69}          & 52.15                   \\
\multicolumn{1}{c|}{4}          & \multicolumn{1}{c|}{CCNet (ours)} & \textbf{91.92} & \textbf{75.1} & \textbf{65.1} & \multicolumn{1}{c|}{\textbf{70.78}} & \multicolumn{1}{c|}{\textbf{75.725}} & \textbf{66.24}        & \textbf{56.71}        & \textbf{93.63}        & \multicolumn{1}{l|}{\textbf{70.36}} & \textbf{71.74}          \\ \bottomrule
\end{tabular}%
}
\label{tab:res}
\end{table*}

\subsection{Datasets}
We evaluate our approach using two widely recognized benchmarks for domain generalization. 

1) \textit{The PACS dataset} \cite{li2017deeper} is specifically curated for domain generalization tasks. It comprises four distinct domains: Photo, Art Painting, Cartoon, and Sketch. This dataset consists of 9,991 images distributed across these domains, with each domain containing seven categories. The dataset's diversity stems from the variety of artistic styles represented, rendering it a challenging benchmark for domain generalization. Figure \ref{fig:pacs_sample} illustrates the challenge.

2) \textit{The VLCS dataset} \cite{fang2013unbiased} is commonly employed for assessing domain generalization methods. It comprises large images categorized into five classes (bird, car, chair, dog, and person), distributed equally across four domains (Caltech101 %\cite{griffin2007caltech}
, LabelMe %\cite{russell2008labelme}
, SUN09 %\cite{choi2010exploiting},
and VOC2007).
%\cite{everingham2010pascal}).
For each benchmark, we conduct leave-one-out testing across all four domains. This entails reserving one domain as the unseen domain in each experiment, training on the remaining three domains, and evaluating the performance on the unseen domain without any fine-tuning.

\subsection{Implementation Details}
We locally train the clients for 5 epochs to ensure local convergence, employing a batch size of 256. Optimization is carried out using the Lion optimizer \cite{chen2024symbolic}, with a learning rate of 0.0003 and weight decay of 0.05. The values of $\beta_1$ and $\beta_2$ are set to 0.95 and 0.98, respectively, as recommended by the authors of Lion to mitigate training instability. We observe that 10 communication rounds between clients are adequate for the convergence of the global model, and this number is fixed for all experiments conducted on all datasets. Our optimal variant of CCNet employs 3 classification heads with identical structures, utilizing the same classification head as in the original Agglomerator.

\subsection{Experiments and Results}
To show the effectiveness of our proposed backbone, we compare it against MobileNet \cite{howard2017mobilenets}, a widely adopted lightweight CNN architecture with a parameter count similar to ours. Although our architecture was not pre-trained on datasets such as ImageNet, it incorporates features from a pre-trained encoder during the initialization step, as detailed in the methodology. Hence, to ensure a fair comparison, we selected a MobileNet architecture pre-trained on ImageNet1k.
The backbone comparison is performed on 5 different FedDG algorithms belonging to different categories.
1) \textbf{\textit{FedAvg}} \cite{mcmahan2017communication} is one of the earliest and well established algorithms in Federated Learning. 2) \textbf{\textit{AM}} \cite{liu2021feddg} is a technique that mixes the frequency information from different clients to increase robustness. 3) \textbf{\textit{FedProx}} \cite{li2020federated} which offers parameterizes the weights for different models to better handle data heterogeneity. 4) \textbf{\textit{RSC}} \cite{huang2020self} which is a regularization based technique 5) \textbf{\textit{Scaffold}} \cite{karimireddy2020scaffold} makes use of data similarity to tackle the client drift. 
Results are shown in Table \ref{tab:algos}. Our backbone using fewer parameters outperforms the CNN backbone on both benchmarks on all the 5 different algorithms.  On PACS, the average performance across the four domains improved by more than 20\% compared to the second-best approach, MobileNet. However, on VLCS, it only led to an average performance improvement of around 4\% compared to MobileNet. It is noteworthy that PACS presents more challenging domain shifts than VLCS, and this is where our proposed approach demonstrates its merits.

\subsection{Ablation Study}
We conduct an analysis that compares our proposed CCNet, and the aforementioned related foundational models that attempt to model the scene parse tree as well as MobileNet. The models are used as backbones for the FedAvg algorithm that acts as a strong baseline for comparison. Results on PACS, and VLCS are shown in Table \ref{tab:res}.

Interestingly, it is observed that the pre-training approach proposed in \cite{radwan2023distilling} performed worse than using the vanilla Agglomerator without any pre-training. This contrasts with the superior performance reported for the same pre-training methodology on a typical supervised classification task. This may suggest that the pre-training solely teaches the model how to build the parse tree with embeddings very specific to the source domain, but fails when transferred to another domain.

We also investigate the impact of varying the number of classification heads in our CCNet model. Specifically, we tested three configurations: 1) Using a single classification head on the last level only, following the approach of Agglomerator and distill part-whole. 2)
Building an extra classification head, one on top of the second level (2 heads). 3) Incorporating one more head on top of the first level (3 heads).
The final classification result is obtained by averaging the predictions from all the heads. The results are summarized in Table \ref{tab:ablation}.
Interestingly, while the last level in the feature representation should encapsulate the most semantic representation of the image, we observed that incorporating classification heads on the parts and sub-parts levels contributed to better generalization and improved accuracy. This highlights the significance of recognizing the structural components of the scene in achieving accurate classification.

\section{Conclusion}
In this paper, we introduce CCNet, a novel approach in the domain of federated domain generalization. This architecture is specifically designed to model the part-whole hierarchies inherent in scenes, which are expected to remain consistent across various domains despite changes in appearance or significant shifts in domain. Our proposed approach showcases robust generalization capabilities, surpassing the performance of both the original Agglomerator model and the distill part-whole pre-training approach without relying on a pre-training step. Additionally, the model surpasses MobileNet, a widely used lightweight CNN architecture, while utilizing fewer parameters and offering the advantage of explainability through the visualization of islands of agreements. We hope that our work will inspire further research in developing models that not only exhibit improved generalizability but also provide enhanced interpretability.

\begin{table}[]
\caption{Ablation on the number of classification heads on top of the different levels for our proposed method.}
\centering
\resizebox{\columnwidth}{!}{
{\renewcommand{\arraystretch}{1.2}
\begin{tabular}{@{}lllllll@{}}
\toprule
\multicolumn{1}{c}{}            & \multicolumn{1}{c}{}                          & \multicolumn{4}{c}{PACS}                                                                                             & \multicolumn{1}{c}{}    \\ \midrule
\multicolumn{1}{c|}{Params (M)} & \multicolumn{1}{c|}{Method}                   & \multicolumn{1}{c}{P} & \multicolumn{1}{c}{\textbf{A}} & \multicolumn{1}{c}{C} & \multicolumn{1}{c|}{S}              & \multicolumn{1}{c}{Avg} \\ \midrule
\multicolumn{1}{l|}{3}          & \multicolumn{1}{l|}{CCNet (1 head)}  & 89.64                 & 72.31                          & 64.33                 & \multicolumn{1}{l|}{63.37}          & 72.4125                 \\
\multicolumn{1}{l|}{3.5}        & \multicolumn{1}{l|}{CCNet (2 heads)} & 87.01                 & 72.51                          & 64.08                 & \multicolumn{1}{l|}{68.9}           & 73.125                  \\
\multicolumn{1}{l|}{4}          & \multicolumn{1}{l|}{CCNet (3 heads)} & \textbf{91.92}        & \textbf{75.1}                  & \textbf{65.1}         & \multicolumn{1}{l|}{\textbf{70.78}} & \textbf{75.725}         \\ \bottomrule
\end{tabular}}}
\label{tab:ablation}
\end{table}

\bibliographystyle{IEEEbib}
\bibliography{refs}

\end{document}